\documentclass[preprint]{elsarticle}

\usepackage{lineno}
\usepackage{xcolor}
\definecolor{HighlightColor}{rgb}{0.69,0.81,0.11} %
\usepackage[pdftex,colorlinks,linkcolor=HighlightColor,citecolor=HighlightColor,filecolor=HighlightColor,urlcolor=HighlightColor,backref,pdfpagelabels,pdfpagemode=UseNone,implicit]{hyperref}
\usepackage{pgfplots}
\usepackage{amsmath}
\usepackage{tikz}
\usepackage{pgfplots}
\usepackage[commentsnumbered,vlined]{algorithm2e}
\pgfplotsset{compat=newest}
\usepgfplotslibrary{polar}
\usepgflibrary{shapes.geometric}
\usetikzlibrary{calc, intersections}
\usetikzlibrary{trees,decorations.pathmorphing}
\definecolor{nvgreen}{RGB}{109,176,76}
\tikzstyle{neuron}=[circle, minimum size=0.5cm, draw=black, text=black,font=\small] 
\tikzstyle{connection}=[->,font=\small, text=black, draw=black]
\tikzstyle{annot} = [text width=4em, text centered, draw=black]
\def\layersep{2.8cm}
\tikzset{
  invisible/.style={opacity=0}, 
  visible on/.style={alt={#1{}{invisible}}},
  alt/.code args={<#1>#2#3}{
    \alt<#1>{\pgfkeysalso{#2}}{\pgfkeysalso{#3}}
  },
}

\graphicspath{{./}{Images/}{../../../12Alias/}}

\journal{Mathematics and Computers in Simulation}

\begin{document}

\begin{frontmatter}

\title{Integral Equations and Machine Learning}

\author{Alexander Keller\fnref{fnkeller}}
\author{Ken Dahm\fnref{fndahm}}
\address{NVIDIA, Fasanenstr. 81, 10623 Berlin, Germany}
\fntext[fnkeller]{keller.alexander@gmail.com}
\fntext[fndahm]{ken.dahm@gmail.com}

\begin{abstract}
As both light transport simulation and reinforcement learning are
ruled by the same Fredholm integral equation of the second kind,
reinforcement learning techniques may be used for photorealistic
image synthesis: Efficiency may be dramatically improved by
guiding light transport paths by an approximate
solution of the integral equation that is learned during rendering.
In the light of the recent advances in reinforcement learning for
playing games, we investigate the representation of an
approximate solution of an integral equation by artificial neural networks
and derive a loss function for that purpose. The resulting
Monte Carlo and quasi-Monte Carlo methods train neural
networks with standard information instead of linear information
and naturally are able to generate an arbitrary number of training samples.
The methods are demonstrated for applications in light transport simulation.
\end{abstract}


\end{frontmatter}

\section{Introduction}

The fast progress in the field of machine learning is becoming increasingly
important for other research areas including Monte Carlo methods and
especially computer graphics. In fact the fields are closely related in
a mathematical sense: Reinforcement learning has been shown
equivalent to solving Fredholm integral equations of the second kind \cite{LightRL}
and is used for simple and efficient importance sampling in light transport simulation.
Furthermore, the process of sampling light transport paths from
path space to simulate light transport has similarities to the process
of computers playing games, as in fact, sampling paths in search trees
amounts to playing random games to learn about winning chances \cite{SurveyMCTS}.

The utility of importance sampling enabled by high-dimensional function
approximation by artificial neural networks \cite{DeepLearning} has recently been demonstrated 
in an impressive way for learning and playing the game of Go \cite{AlphaZero},
bringing together the aforementioned domains even closer.

We therefore briefly review the equivalence of
reinforcement learning and Fredholm integral equations
of the second kind in Sec.~\ref{Sec:LightRL} and point
out further analogies to recent advances in machine learning
for playing games in Sec.~\ref{Sec:AlphaZero}.
We then derive a scheme to train artificial neural networks
within integral equations in Sec.~\ref{Sec:Train} and explore
applications in computer graphics:
In Sec.~\ref{Sec:TDL} a combination of temporal difference
learning and artificial neural networks is used for importance
sampling light sources and Sec.~\ref{Sec:Baking} 
investigates the feasibility of high-dimensional
approximation by artificial neural networks for real-time
rendering.

\section{Importance Sampling by Reinforcement Learning} \label{Sec:LightRL}

Physically based rendering \cite{PBRT} in principle consists of
summing up the contributions  of light transport paths that connect
the camera with the light sources. Due to the large state space, finding 
contributing paths may be inefficient, for example, because
visibility needs to be sampled and is not known up front.

Using reinforcement learning to learn where light is coming
from \cite{LightRL} allows for efficient importance sampling of light transport
paths. The method has been derived by  matching terms of
the Q-learning \cite{Qlearning} and light transport equations:
\begin{eqnarray}
		Q'(s, a)      &= &(1 - \alpha) \cdot Q(s, a) + \ \alpha 
			\cdot \bigg( r(s, a)
			+ {\gamma}{\int_{\mathcal A}}
			\pi(s', a')
			Q(s', a')
			da' \bigg) \nonumber \\
		L(x, \omega) &=
			& \qquad L_e(x, \omega)
			+ \int_{{\mathcal S}^2_+(x)}
			f_s(\omega_i, x, \omega) \cos \theta_i
			L(h(x, \omega_i), -\omega_i)
			d\omega_i \label{Eqn:Fredholm}
\end{eqnarray}
In Q-learning, the value $Q(s, a)$ of taking an action $a$ in
state $s$ is learned by taking the fraction $1- \alpha$ of the
current value and adding the fraction $\alpha$ of the reward
$r(s,a)$ received when taking action $a$. In addition, all future reward
is discounted by a factor of $\gamma$. It is determined
by the integral over all actions $a'$
that can be taken from the next state $s'$ that is reached from
state $s$ by taking action $a$ is computed over the
values $Q(s', a')$ weighted by a so-called policy function $\pi(s', a')$.

While this may sound abstract at first, matching the terms
with the integral equation describing radiance transport
immediately shows the parallels between reinforcement
learning and light transport: In fact the reward corresponds
the radiance $L_e$ emitted by the light sources, the
discount factor times the policy function corresponds to
the reflection properties $f_s$ (also called the bidirectional
scattering distribution function), telling how much radiance
is transported from direction $\omega_i$ to direction $\omega_r$
through the point on a surface $x$. The value $Q$ then
can be matched with the radiance $L$ that comes from the
point $h(x, \omega_i)$ hit by tracing a ray along a straight
line from $x$ into direction $\omega_i$ and the state space
$\mathcal A$ corresponds to the hemisphere ${\mathcal S}^2_+(x)$
aligned by the normal in point $x$. This leaves us with
an action $a$ corresponding to tracing a ray.

Combining both equations yields
 \begin{eqnarray} \label{Eqn:QLearning}
  Q'(x, \omega) & = & (1 - \alpha) Q(x, \omega) \\
  &&+ \alpha \left(L_e(y, -\omega) + \int_{{\mathcal S}^2_+(y)} f_s(\omega_i, y, -\omega) \cos \theta_i Q(y, \omega_i ) d\omega_i\right) \nonumber
\end{eqnarray}
where $Q$ now represents the incident radiance, which
can be learned over time and be used for importance
sampling directions proportional to where most light is coming from,
i.e. guiding path towards the light sources.
The implementation of such algorithms is detailed in \cite{LightRL}.

Besides the structural identity of reinforcement learning and a Fredholm
integral equation of the second kind, there are more analogies:
While $Q$-learning considers the value of the next, non-terminal state,
which corresponds to scattering in transport simulation \cite{LightRL},
temporal difference learning \cite{TDlearning} is related to next
event estimation, and as such deals with terminal states
as discussed in Sec.~\ref{Sec:TDL}.

\begin{figure}
	\centering
	\begin{tikzpicture}
	[level distance=15mm,
	level 1/.style={sibling distance=12mm},
	every node/.style={gray,thick,inner sep=2pt,font=\small},
	every path/.style={gray,thick},
	intree/.style={draw,circle,nvgreen,thick,font=\small,minimum size=0.9cm,text=black},
	scale=1 ]

	\node [intree] (ca) at (7,-1) {root}
		child {node (cb) [intree] {$\frac{0}{1}$}
		}
		child [missing]
		child {node (cc) [intree] {$\frac{5}{17}$}
			child {node (cf) [intree] {$\frac{2}{8}$}}
			child {node (cg) [intree] {$\frac{3}{9}$}}
		}
		child [missing]
		child {node (cc) [intree] {$\frac{1}{6}$}
		}
		child [missing]
		child {node (cc) [intree] {$\frac{13}{18}$}
			child {node (cf) [intree] {$\frac{6}{7}$}}
			child {node (cg) [intree] {$\frac{4}{5}$}}
			child {node (cg) [intree] {$\frac{3}{6}$}}
	};
	\node (cl) [black, below of=cf,yshift=-9mm] {lose $0/1$ win};
	\draw[decorate,decoration=zigzag,->,gray] (cf)--(cl);

	\draw[black, latex-, shorten >= 15, shorten <= 15]  ($(cc)+(-0.1, 0.1)$) -- ($(cf)+(-0.1, 0.1)$);
	\draw[black, latex-, shorten >= 15, shorten <= 15]  ($(ca)+(-0.1, -0.1)$) -- ($(cc)+(-0.1, -0.1)$);
\end{tikzpicture}
\caption{Illustration of the principle of Monte Carlo tree search to find the most valued move represented by the
branches from the root node: All children contain a fraction, where the denominator
counts the number of visits to the node and the nominator is the reward.
In order to update the
values, a move is selected considering the values of the children of a
node until a leaf node is reached. Unless this leaf node is expanded (see main
text), random moves are taken (zig-zag line) until a terminal state of the game is
reached. Then the nodes along the path back to the root (along the arrows) are updated
by incrementing the number of visits and also incrementing the reward
unless the random game has been lost.
\label{Fig:MCTS}}
\end{figure}
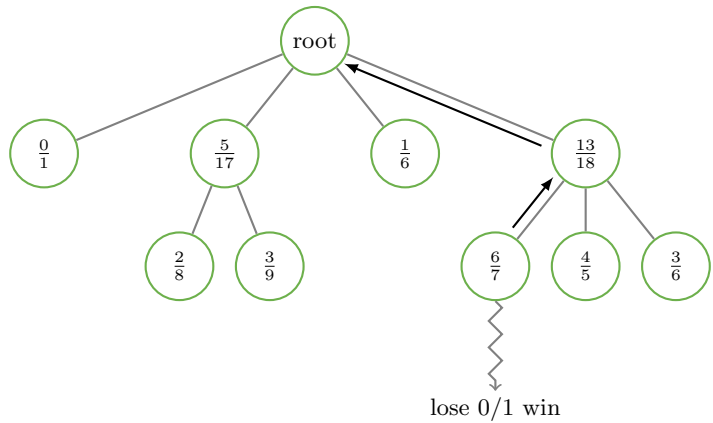

\section{Importance Sampling in Monte Carlo Tree Search} \label{Sec:AlphaZero}

When a computer program needs to decide, which next move in a game
is best, it does so by computing a value for each possible move and then
uses a heuristic to select the best one. Implementing the rules of a
game, it is straightforward to describe the tree, where each path corresponds to
the sequence of moves of one complete game. As for most interesting games
this tree is growing exponentially with the number of moves,
building such a tree is not feasible for most games and hence various tree search
heuristics have been explored.

The currently most powerful algorithms \cite{AlphaZero} are based on Monte Carlo
tree search as illustrated in Fig.~\ref{Fig:MCTS}. Instead of trying to
evaluate the complete search tree, the key principle is to estimate
the value of a move by randomly sampling a path in the search tree
and counting how often such random games are won. Repeating the
procedure the statistics can be improved, while the running time stays
linear in the maximal number of possible moves of the games. 
Note that randomly selecting moves in a game resembles selecting random
scattering directions in light transport simulation to generate light
transport paths.

In order to increase the efficiency of Monte Carlo tree search,
parts of the search tree may be stored as illustrated in Fig.~\ref{Fig:MCTS}.
Nodes then store their total number of visits $n$ and the reward $w$,
when passing through them. Whenever a node is reached
during tree traversal that is not the terminal node of a game,
children may be appended to count results of the random tree exploration.
While the number of stored nodes is bounded by available memory,
there are additional heuristics to only create new nodes once their
parents have a sufficiently high value and number of visits.

\subsection{Action Selection and Simulation of Densities}

When selecting an action, i.e. a move in the game, exploration
and exploitation are competing goals. On the one hand, the
tree search needs to be able to simulate all possible games
in order to learn. On the other hand, game play should be
strong and highly valued moves should be preferred. This balance
is achieved by computing the so-called upper confidence
bound \cite{BanditbasedMCP}
\begin{eqnarray*}
	u = \frac{w}{n} + c \cdot \sqrt{\frac{\ln N}{n}} , %
\end{eqnarray*}
for each possible move for the current state. The
value $\frac{w}{n}$ of games won passing through the node
under consideration accounts for exploitation, as it is
large for likely good moves. With $N$ being the total number
of simulated games, the second term remains large
for nodes that have not been visited often. With the
constant $c$ usually set to $\sqrt{2}$, this term ensures
exploration. As long as all values are available, the next move
is selected as the child with maximum $u$ value during Monte
Carlo tree search.

This method overcomes previous methods like,
for example, $\epsilon$-greedy action selection,
which would just choose the child with maximal $\frac{w}{n}$,
unless a random number is less than a threshold $\epsilon$, in which case
a random action would be taken.
As it is very close to what can be done in reinforcement learned
importance sampling in light transport simulation, we mention
probabilistic action selection, which assigns the probability
\begin{equation} \label{Eqn:ProbSelect}
  \text{Prob}(a_i \mid s) = \frac{T^{Q(s, a_i)}}{\sum_{a_k}T^{Q(s, a_k)}}
\end{equation}
to each possible action $a_i$. Then a
small $T$ favors exploration, while large $T$ tend to exploitation.
Starting with a small $T$ during learning, the probabilities remain
more uniform, while with growing $T$ actions with higher value
will become selected as the values $Q$ become more reliable.

Action selection in reinforcement learned importance sampling
in light transport simulation amounts to
sampling a direction proportional to $Q$ (see Eqn.~\ref{Eqn:QLearning})
rather than selecting the maximum of a distribution.

Guaranteeing that exploration always remains possible corresponds
to guaranteeing ergodicity in transport simulation. Hence
all densities and values are required to be non-zero as long as
there may be a non-zero contribution of a path.

\subsection{Artificial Neural Networks for Densities and Values}

While random play is feasible, it is not really efficient when a large
number of samples is required in order to achieve reliable values
for action selection. Especially for large state or action spaces
this may become a bottleneck. Using expert knowledge can
improve the selection process \cite{GoogleGo}, but requires
the acquisition and formalization of such knowledge.

Reinforcement learning and self-play \cite{TDGammon,AlphaZero}
have been demonstrated to overcome these issues.
In analogy to reinforcement learned importance sampling
in light transport simulation \cite{LightRL}, key to efficiency is guiding
the paths in the search tree along the most rewarding nodes
even during random play, where no information has been stored, yet.
As discrete representations are not feasible due to the size of
the state space alone,
deep artificial neural networks \cite{DeepLearning} have been trained to predict the values of moves.
In Sec.~\ref{Sec:TDL} and Sec.~\ref{Sec:Baking}, we discuss
artificial neural networks to replace discrete representations
in light transport simulation.

\section{Training Artificial Neural Networks within Integral Equations} \label{Sec:Train} %

The artificial neural networks in \cite{AlphaZero} are trained by self-play,
where reinforcement learning is realized by having the computer
play against itself. This is possible by Monte Carlo tree search
exploring random games according to the rules of the game
as described in Sec.~\ref{Sec:AlphaZero}.

Now light transport is ruled by the Fredholm integral equation~(\ref{Eqn:Fredholm})
and training an artificial neural network requires an error, a so-called loss function.
This error is derived by taking a look at  Eqn.~(\ref{Eqn:QLearning}): Assuming
that learning has converged, meaning $Q' = Q$, yields $\alpha = 1$ and hence %
\begin{eqnarray*}
   Q(x, \omega) = L_e(y, -\omega) + \int_{{\mathcal S}^2_+(y)} f_r(\omega_i, y, -\omega) \cos \theta_i Q(y, \omega_i ) d\omega_i .
\end{eqnarray*}
Representing $Q(x, \omega)$ by an artificial neural network $\hat Q(x, \omega)$ \cite{DeepLearning},
lends itself to defining the error
\begin{eqnarray*}
  \Delta Q(x, \omega) := \hat Q(x, \omega) - \left(L_e(y, -\omega) + \int_{{\mathcal S}^2_+(y)} f_r(\omega_i, y, -\omega) \cos \theta_i \hat Q(y, \omega_i ) d\omega_i\right)
\end{eqnarray*}
as the difference between the current value of the network $\hat Q(x, \omega)$
and the more precise value as evaluated by the term in brackets. During learning
$\Delta Q(x, \omega)$ is used to train the artificial neural
network $\hat Q(x, \omega)$ by back-propagation \cite{Back-Propagation}.

An online quasi-Monte Carlo algorithm for training the
artificial neural network $\hat Q(x, \omega)$ by reinforcement learning then
\begin{itemize}
\item generates light transport paths using a low discrepancy sequence \cite{NutshellQMC} and
\item for each vertex of a path evaluates $\Delta Q(x, n)$ to
\item train the artificial neural network $\hat Q(x, n)$ by back-propagation.
\end{itemize}
Other than the classic training of artificial neural networks, the training
set is infinite and in fact each generated path is unique an
used for training exactly once. As any number of training
samples can be generated by a deterministic low discrepancy
sequence, the samples are perfectly reproducible, which allows
for efficiently exploring the hyperparameters of the artificial neural
network. This method falls in the same class as methods
training artificial neural networks without clean data \cite{Noise2Noise}.
In fact standard information in the form of samples is
used instead of using linear information \cite{TWW:88} in the
form of functionals of the solution of the integral equation.

\section{Learning Next Event Estimation} \label{Sec:TDL}

Recent research \cite{Rainbow,AtariDeepRL} has shown that deep
artificial neural networks \cite{DeepLearning} very successfully can approximate
value and policy functions in temporal difference learning or reinforcement 
learning.
In physically based rendering, for example path tracing, we need to find
efficient policies for selecting good scattering directions or light
sources for next event estimation that have high importance.

In order to compute the direct illumination, we have to integrate the
radiance $L_i$ over the surfaces of all light sources, i.e.
\begin{eqnarray} \label{Eqn:NEEInt}
	L(x, \omega) & =  & \int_{y \in \text{supp} L_i} L_i(\omega_i, y) f_r(\omega, x, \omega_i) G(x, y) V(x, y) dA_y \\
 \label{Eqn:NEE}
	& \approx & \frac{1}{N}\sum_{i=0}^{N-1} \frac{L_i(\omega_i, y_i) f_r(\omega, x, \omega_i) G(x, y_i) V(x, y_i)}{p(y_i, \omega_i)} , 
\end{eqnarray}
which is estimated by the Monte Carlo method, as
in practical settings there is no analytical solution.
In order to reduce variance, one of the main challenges is
picking the probability density function $p(y_i, \omega_i)$ for
light source selection. The obviously best choice is a function
proportional to the integrand, however, this is not an option as the
visibility term $V$ is expensive to evaluate and discontinuous
in nature. We therefore utilize temporal-difference
learning \cite{ReinforcementLearning} to learn the distribution over time, i.e.
\begin{equation*}	
	Q'(s, i) = (1 - \alpha)Q(s, i) + \alpha L_i(\omega_i, y_i) f_r(\omega, x, \omega_i) G(x, y_i) V(x, y_i) ,
\end{equation*}
where $s$ is the current state, i.e. a discretized location, normal vector,
and incoming direction, and $i$ is the light source 
index. This gives us a temporal average of all contributions
we have evaluated so far for certain locations in space. The original
temporal difference algorithm discretizes the state and action space
in order to compute these averages.

As compared to previous approaches, approximating the function $Q(s, i)$ by an artificial neural network may
avoid discretization artifacts and use less memory to store the $Q(s, i)$
values.

\subsection{Artificial Neural Network Training and Rendering} \label{Sec:OnlineTraining}

Fig.~\ref{Alg:Rendertraining} outlines the algorithm, which simultaneously
trains an artificial neural network and renders the image.
Operating in mini-batches of $M = 64$ quasi-random samples, a ray is traced
from the eye through each pixel identified by a sample of the
Halton sequence \cite{Hal:64} and intersected with the scene geometry
to find the closest point of intersection. At this point, a light source
for direct illumination is selected and its contribution is computed
according to Eqn.~(\ref{Eqn:NEE}).

The light source is selected using an artificial neural network, whose input layer
consists of nine neurons, receiving the normalized position, normal
vector, and incoming direction of the intersection point. The output
layer contains as many neurons as there are light sources and provides
a $Q$ value for each light source and is realized by the so-called
"softmax" activation function \cite{DeepLearning}. All other layers in the network use
rectified linear units (ReLU). Finally, the light source then is selected
by sampling using the cumulative distribution function.

The contribution of a light source sample as well as the intersection
details are stored in an array and once $M$ mini-batches are complete,
we retrain the network using stochastic gradient descent \cite{Back-Propagation}.

\begin{figure}
	\begin{algorithm}[H]
		\DontPrintSemicolon
		\SetKwProg{Fn}{Function}{}{}
		\SetKwProg{ElseIf}{else if}{}{}
		\Fn{renderImage()}{
			initNetwork()\;
			\For{$i = 1$ \KwTo $N$}{
				\For{$j = 1$ \KwTo $M$}{
					\For{$k = 1$ \KwTo $64$}{
						$p$ $\gets$ selectPixel($\xi_0^{i,j,k}, \xi_1^{i,j,k}$)\;
						$x, n, r$ $\gets$ traceRay()\;
						$qs$ $\gets$ neuralNetwork($x, n, r$)\;
						$cdf$ $\gets$ buildCdf($qs$)\;
						$l$ $\gets$ sampleCdf($cdf, \xi_2^{i,j,k}$)\;
						$c$ $\gets$ getContribution($l$)\;
						addContributionToImage($p, c$)\;
						minibatches$[j][k]$ $\gets$ $x, n, r, l, c$\;
					}
				}
				retrainNetwork(minibatches)\;
			}
			outputImage()\;
		}
	\end{algorithm}
\caption{Pseudocode for rendering direct illumination: An artificial neural network
is trained in mini batches during rendering so that over time the
light source will be selected proportional to Eqn.~(\ref{Eqn:NEEInt}).
\label{Alg:Rendertraining}}
\end{figure}

\subsection{Results}

\begin{figure}
    \centering
	\includegraphics[width=0.46\linewidth]{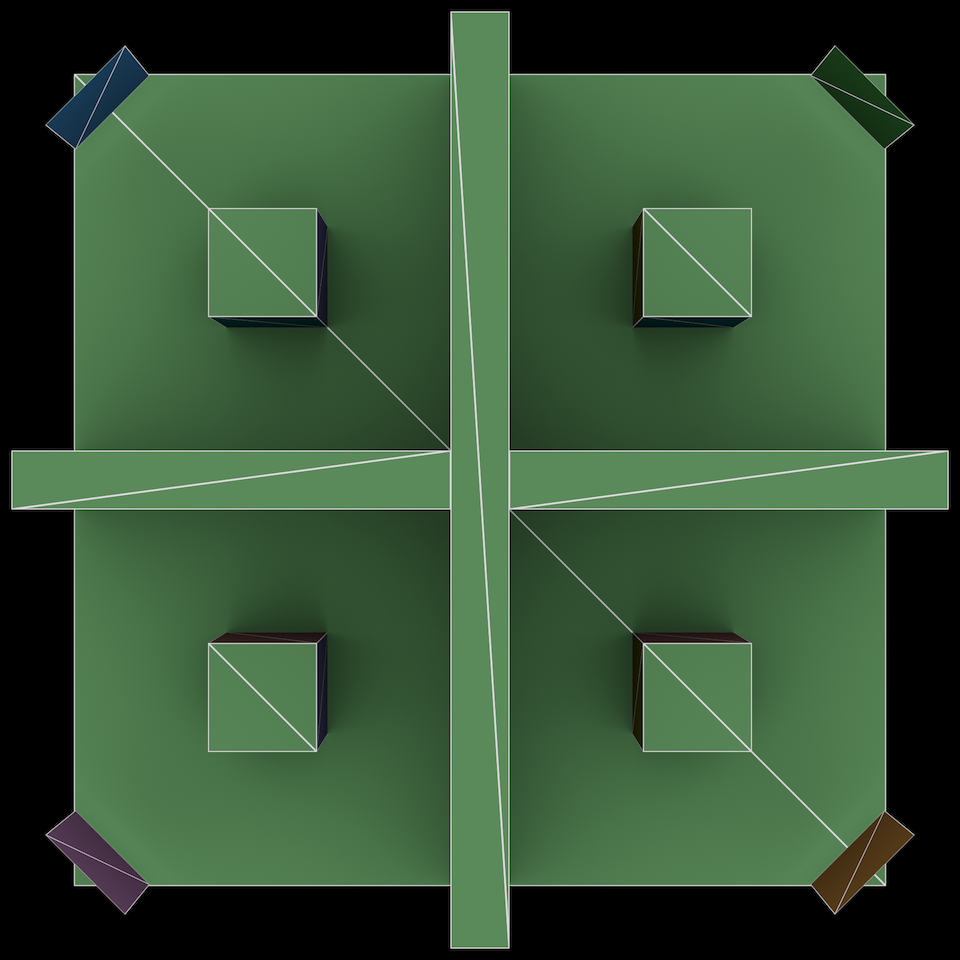}
	\includegraphics[width=0.46\linewidth]{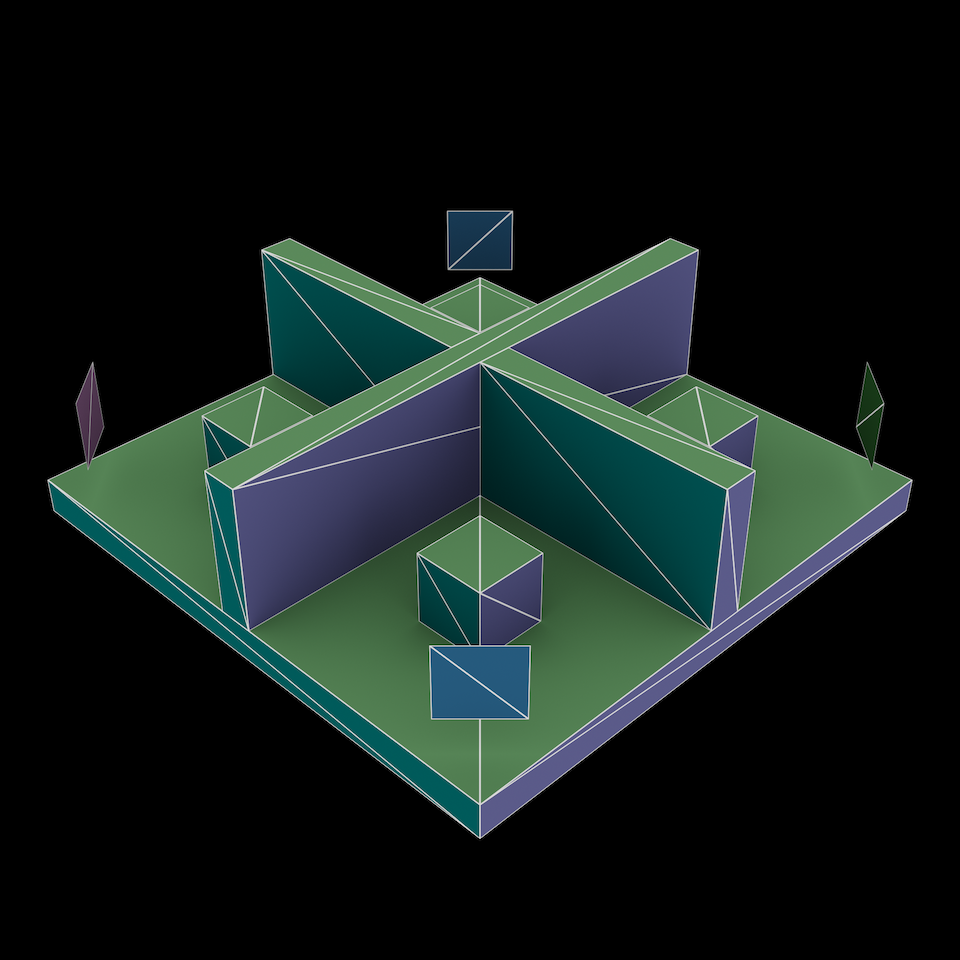}

	\includegraphics[width=0.46\linewidth]{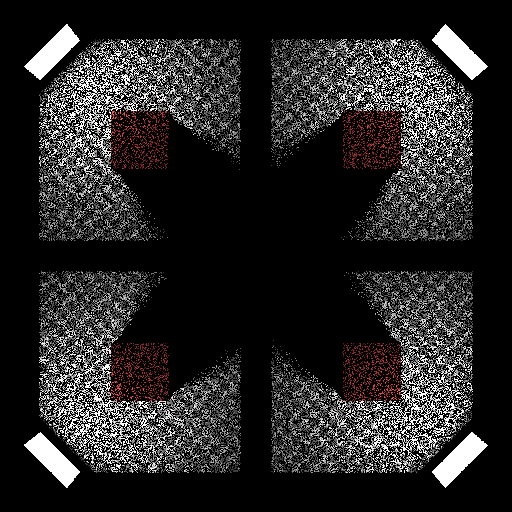} %
	\includegraphics[width=0.46\linewidth]{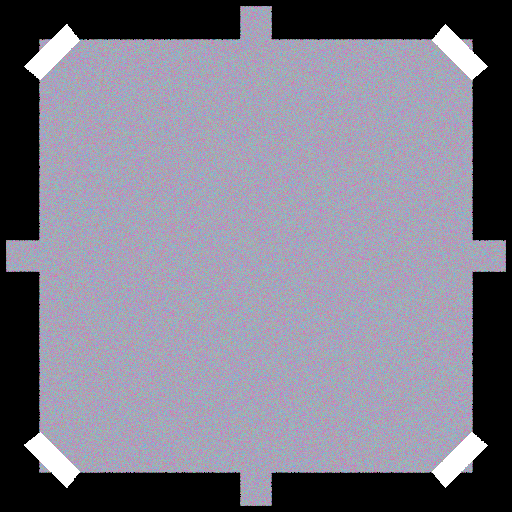} %
	
	\includegraphics[width=0.46\linewidth]{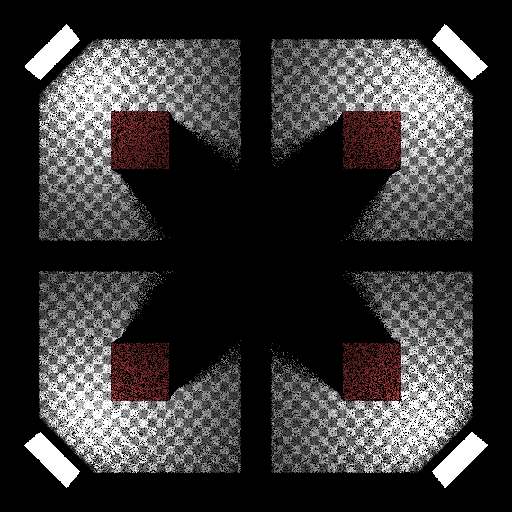} %
	\includegraphics[width=0.46\linewidth]{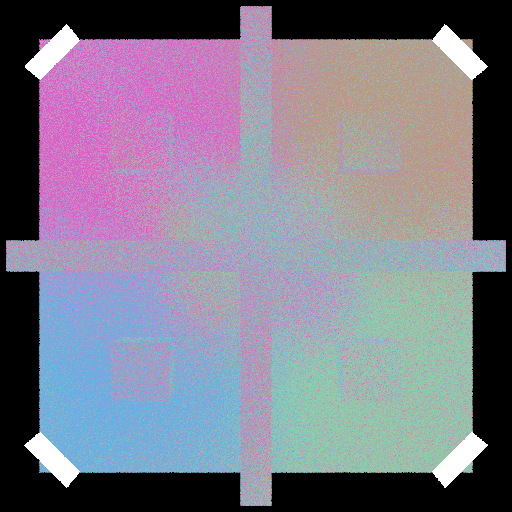} %

    \caption{Top row: Geometry of a test scene with eight light sources in four independent rooms, where in each room only
	two light sources are visible. The other images on the left show the results of uniform light source selection (middle), whereas the bottom image shows the result using an artificial neural network learning the importance of a light source including visibility. The false color visualization on the right illustrates the index of the light
	source selected, where the color noise in the middle shows uniformly random selection, while at the bottom the clear
	coloring illustrates that with high probability only visible light sources are picked.\label{Fig:Split_room}}
\end{figure}

Fig.~\ref{Fig:Split_room} shows the results of an exploratory experiment,
where a small artificial neural network with 2 hidden layers of 64 neural units each
has been used to learn the light source contribution including visibility.
The reduction in variance as compared to uniform light source selection
is clearly visible.

\subsection{Discussion}

The artificial neural network to approximate the importance of a light source including visibility
can be trained during rendering or ahead of image synthesis. While the gain in
efficiency has been demonstrated in an experiment, there are two major challenges
that need to be solved: First, evaluating an artificial neural network per thread on a parallel processor
is much more expensive than looking up a value of importance in a table. 
The second issue is the costly computation of that cumulative distribution function.

Note the when combining next event estimation with path tracing with reinforcement
learned importance sampling \cite{LightRL} by multiple importance sampling
\cite{Veach95-OCSTM,VeachPhD} of course the integrands
weighted by the multiple importance sampling weights need to be learned
instead of the unweighted ones.

\section{Learning Radiance} \label{Sec:Baking}

In computer games lighting has to be computed in real-time. Yet, given
the compute power available in a typical game console, physically accurate
lighting may only be approximated. 
Approximating the radiance $L(x, \omega)$ had not been very
practical, since the function is not smooth and depends on at least five dimensions,
i.e. the three-dimensional location $x$ in space and the direction of observation
$\omega$.

Computing and storing parts of the
illumination information ahead of game play is common practice.
For example, often only the incident indirect radiance is
stored for discrete locations $x$: The radiance
incident in such a location has been approximated using
spherical harmonics \cite{EnvMapSpherical} and
wavelets \cite{WaveletShadows}, although the radiance
incident over the sphere usually is neither smooth nor
piecewise constant. While such representations are
computationally efficient, they require a considerable
amount of memory for the coefficients across the locations
$x$. Adaptive data structures \cite{Lafortune95-FTRVM,PracticalGuiding} are more
suitable for offline rendering, as their construction
is more costly and involved.

In a similar way, artificial neural networks have been explored
to address the curse of dimension in computer graphics:
In \cite[see Fig.2 and first paragraph in App.A]{GI-ANN}, a
4-layer artificial neural network with 20 hidden neurons in
the 2nd and 10 hidden neurons in the 3rd layer has been used to approximate
the red, green, and blue components of the indirect radiance given position $x$,
viewing direction $\omega$, both the
surface normal $n$ and texture information $a$ in $x$, as well as
the position $l_k$ of the $k$-th point light source.
The radiance of each point light source is determined by evaluating
the artificial neural network multiplied by the point light source color $c_k$
and summed up to approximate the indirect radiance in $x$.

In \cite{SkyANN}, the temporally dependent, rather smooth radiance function of a sky
illumination model on the sphere has been successfully approximated by
a small artificial neural network. In contrast, using one tiny artificial neural network to
control the computational cost of the approximation indirect radiance in a complex scene did not work
out in \cite{GI-ANN}:  Decreasing the approximation error
required to adaptively partition the scene hierarchically using a 3d-tree, 
where each voxel refers to one artificial neural network.
While the evaluation of indirect radiance 
is close to real-time, finding such a partition and training the
artificial neural networks remains very computationally expensive. 

\subsection{Baking Radiance into an Artificial Neural Network} \label{Sec:BakingNN}

In order to explore how well tiny artificial neural networks can approximate
the solution $L(x, \omega)$ of the radiance integral equation (\ref{Eqn:Fredholm})
and given the findings of \cite{GI-ANN} as summarized above,
the scene is partitioned into a grid of uniform voxels. In a first step, the surface of the scene is uniformly
sampled and for each point on the scene surface the incoming radiance
is computed for a selected random direction on the unit hemisphere aligned
by the surface normal.
The sample's position in space with its surface normal and the incoming
direction as well as the computed radiance value are stored in a training data list associated
with the voxel the sample lies in.
Then the second step of the algorithm trains a fully connected artificial neural network
for each voxel using the stored training data.
Finally, given a point in the scene, we approximate its
radiance by identifying its voxel
and evaluating its associated artificial neural network given  
the position, normal vector, and incoming direction. In contrast to \cite{GI-ANN} we train the 
artificial neural network on the full light transport equation, meaning we approximate direct and
indirect illumination at once instead of computing direct illumination separately.

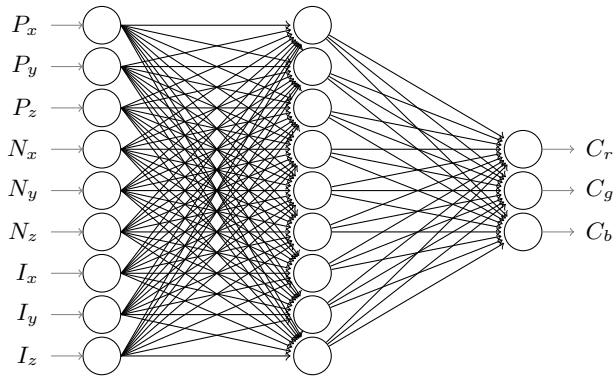
\begin{figure}[t!]
	\centering
	\begin{tikzpicture}[shorten >=1pt,->,draw=black, node distance=\layersep]
		\path node[neuron, pin={[pin edge={<-}]left:\small $P_x$}] (I-1) at (0,-1 * 0.55) {};
		\path node[neuron, pin={[pin edge={<-}]left:\small $P_y$}] (I-2) at (0,-2 * 0.55) {};
		\path node[neuron, pin={[pin edge={<-}]left:\small $P_z$}] (I-3) at (0,-3 * 0.55) {};
		\path node[neuron, pin={[pin edge={<-}]left:\small $N_x$}] (I-4) at (0,-4 * 0.55) {};
		\path node[neuron, pin={[pin edge={<-}]left:\small $N_y$}] (I-5) at (0,-5 * 0.55) {};
		\path node[neuron, pin={[pin edge={<-}]left:\small $N_z$}] (I-6) at (0,-6 * 0.55) {};
		\path node[neuron, pin={[pin edge={<-}]left:\small $I_x$}] (I-7) at (0,-7 * 0.55) {};
		\path node[neuron, pin={[pin edge={<-}]left:\small $I_y$}] (I-8) at (0,-8 * 0.55) {};
		\path node[neuron, pin={[pin edge={<-}]left:\small $I_z$}] (I-9) at (0,-9 * 0.55) {};

		\foreach \name / \y in {1,...,9} 
 			\path node[neuron] (Ha-\name) at (\layersep,-\y * 0.55) {}; %

		\path node[neuron, pin={[pin edge={->}]right:\small $C_r$}] (O-1) at (2 * \layersep,-4 * 0.55) {};
		\path node[neuron, pin={[pin edge={->}]right:\small $C_g$}] (O-2) at (2 * \layersep,-5 * 0.55) {};
		\path node[neuron, pin={[pin edge={->}]right:\small $C_b$}] (O-3) at (2 * \layersep,-6 * 0.55) {};

		\foreach \source in {1,...,9}
			\foreach \dest in {1,...,9}
				\draw[connection, ->] (I-\source.east) -- (Ha-\dest);

		\foreach \source in {1,...,9}
			\foreach \dest in {1,...,3}
				\path[connection] (Ha-\source) edge (O-\dest);
	\end{tikzpicture}  
\caption{Example of a fully connected artificial neural network as used for the
radiance representation inside a voxel. All activation functions
are rectified linear units (ReLU). Given position $P$, surface normal $N$,
and incident direction $I$, the color $C$ is determined.
\label{Fig:net_arch}}
\end{figure}

Fig.~\ref{Fig:net_arch} shows the artificial neural network architecture that is used
inside each voxel. It consists of an input layer
of nine neurons (i.e. normalized position inside the voxel, normal
vector, incoming direction), one hidden layer of nine neurons and
an output layer of three neurons for the radiance in the RGB color space.
In accordance with \cite{GI-ANN}, Experiments indicate that the more complex the
geometry, the more layers and/or neurons are required in order to
approximate the radiance $L(x, \omega)$. Hence keeping the artificial neural
networks small requires a finer partition of the scene in
such situations. In addition, encoding directions in Cartesian coordinates turned out to be
more efficient than using spherical coordinates. 

\subsection{Results}

\begin{figure}[t!]
	\centering
	\includegraphics[width=0.32\linewidth]{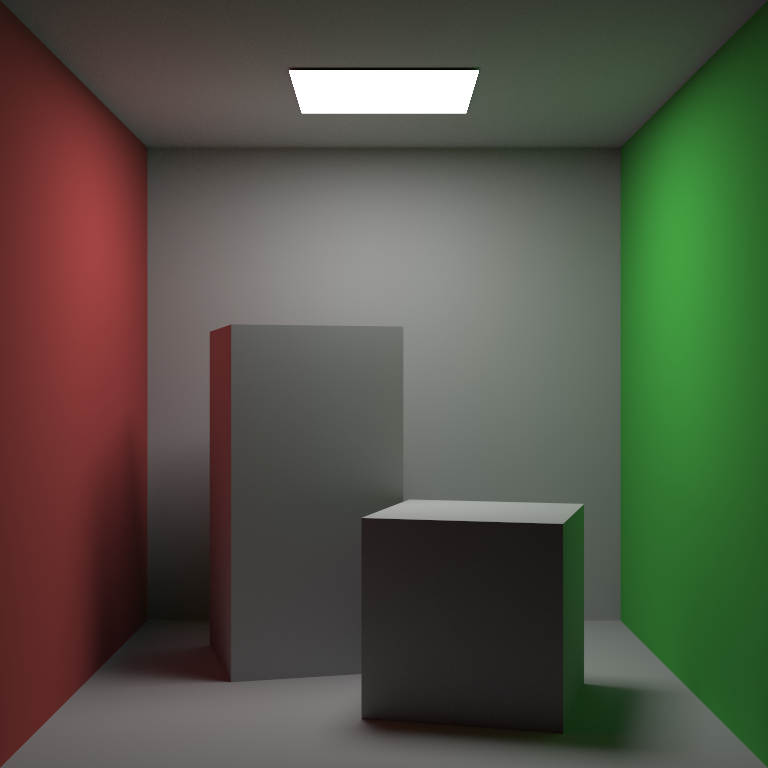}
	\includegraphics[width=0.32\linewidth]{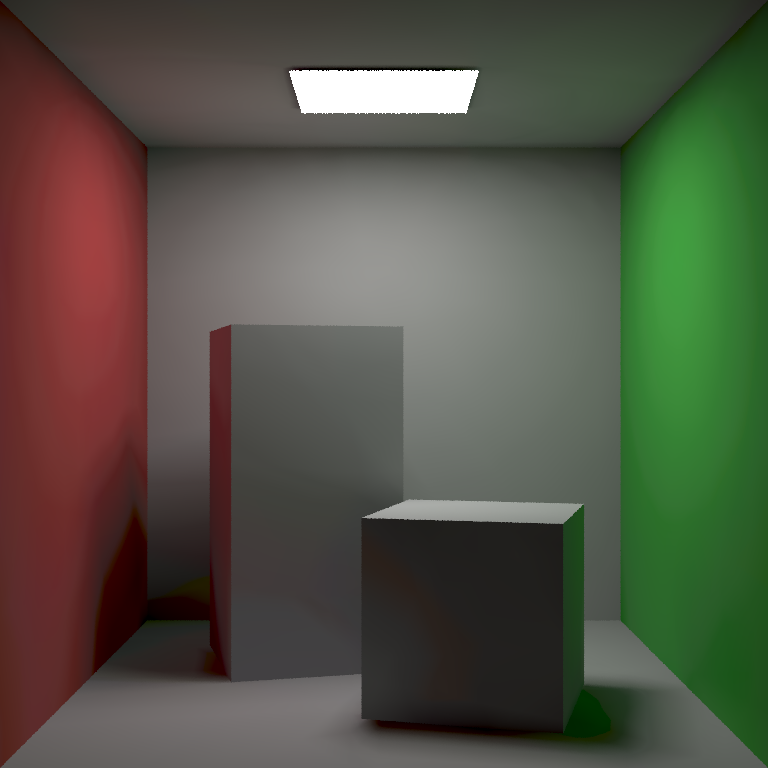}
	\includegraphics[width=0.32\linewidth]{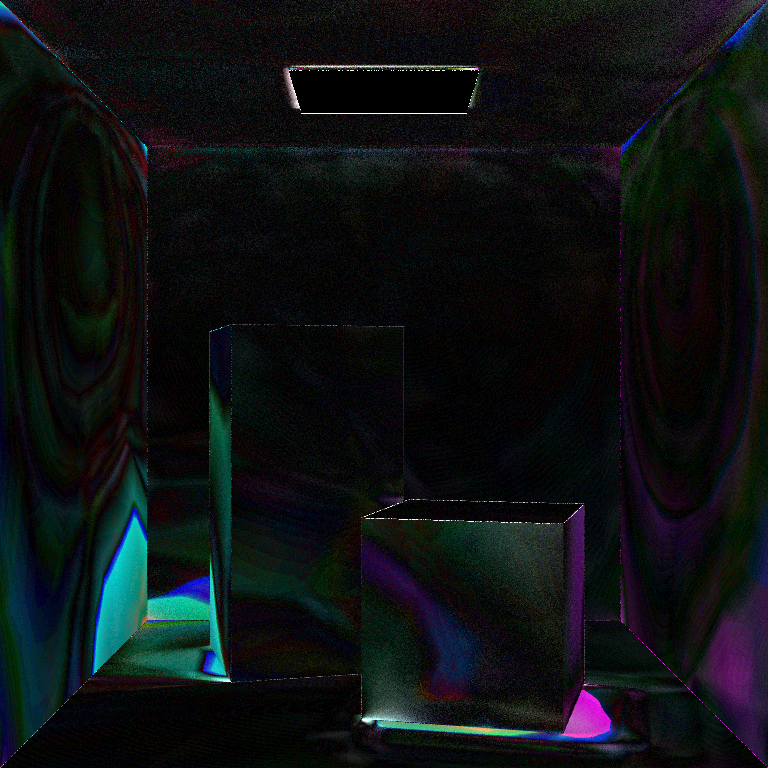}
	\includegraphics[width=0.32\linewidth]{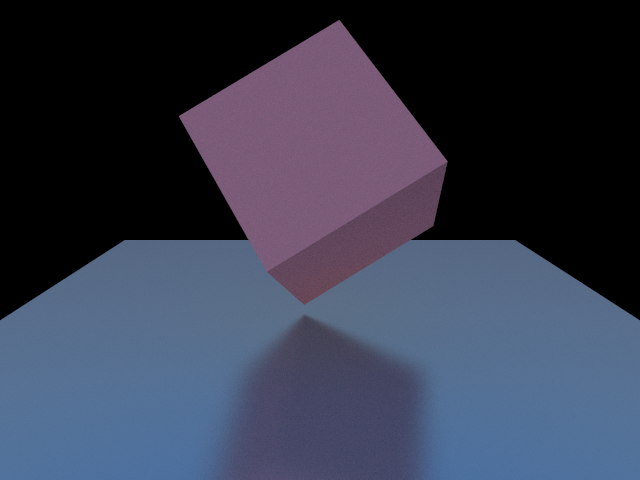}
	\includegraphics[width=0.32\linewidth]{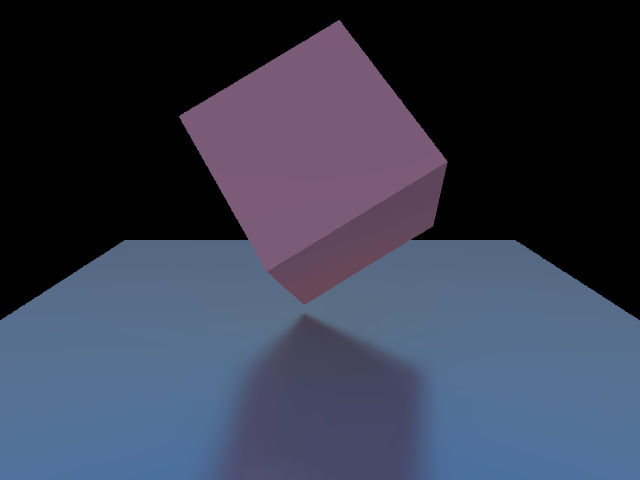}
	\includegraphics[width=0.32\linewidth]{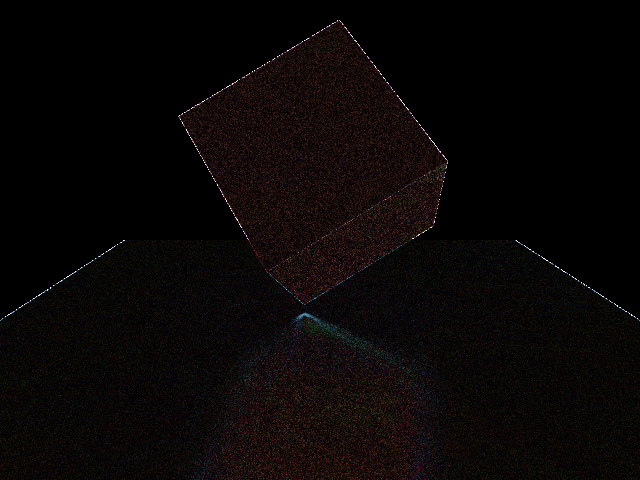}
\caption{Approximation of the radiance $L(x, \omega)$ by artificial neural networks: The
left column has been rendered by path tracing at 512 paths per pixel,
while the middle column shows the approximation by artificial neural networks,
and the right column visualizes their squared difference amplified by a
factor of 10. The top row shows the diffuse
Cornell box, while the bottom row features glossy reflections.
\label{Fig:NeuralApprox}}
\end{figure}

Fig.~\ref{Fig:NeuralApprox} compares the approximation
by artificial neural networks to a
reference image by showing the squared difference. 
The reference has been path traced
with 512 paths per pixel of length six. The scenes have been
discretized into $3 \times 3 \times 3$ voxels requiring 27 artificial neural networks.
10000 points have been distributed uniformly across the scene
and each point has been inserted into its respective voxel together
with the computed radiance values after shooting 512 random
rays into its local hemisphere. While the approximation
is already good at one sample per pixel, 4 samples
per pixel were used for anti-aliasing in order to attenuate the approximation
error along edges (see the cube silhouette in the error
image in Fig.~\ref{Fig:NeuralApprox}). The error images
were scaled by a factor of 10, since otherwise differences
would be hard to spot qualitatively.
In comparison to traditional light baking techniques,
view-dependent effects such as glossy reflections
can be efficiently approximated by artificial neural networks,
which is their most prominent advantage.

\begin{figure}[t!]
	\centering
	\includegraphics[width=0.32\linewidth]{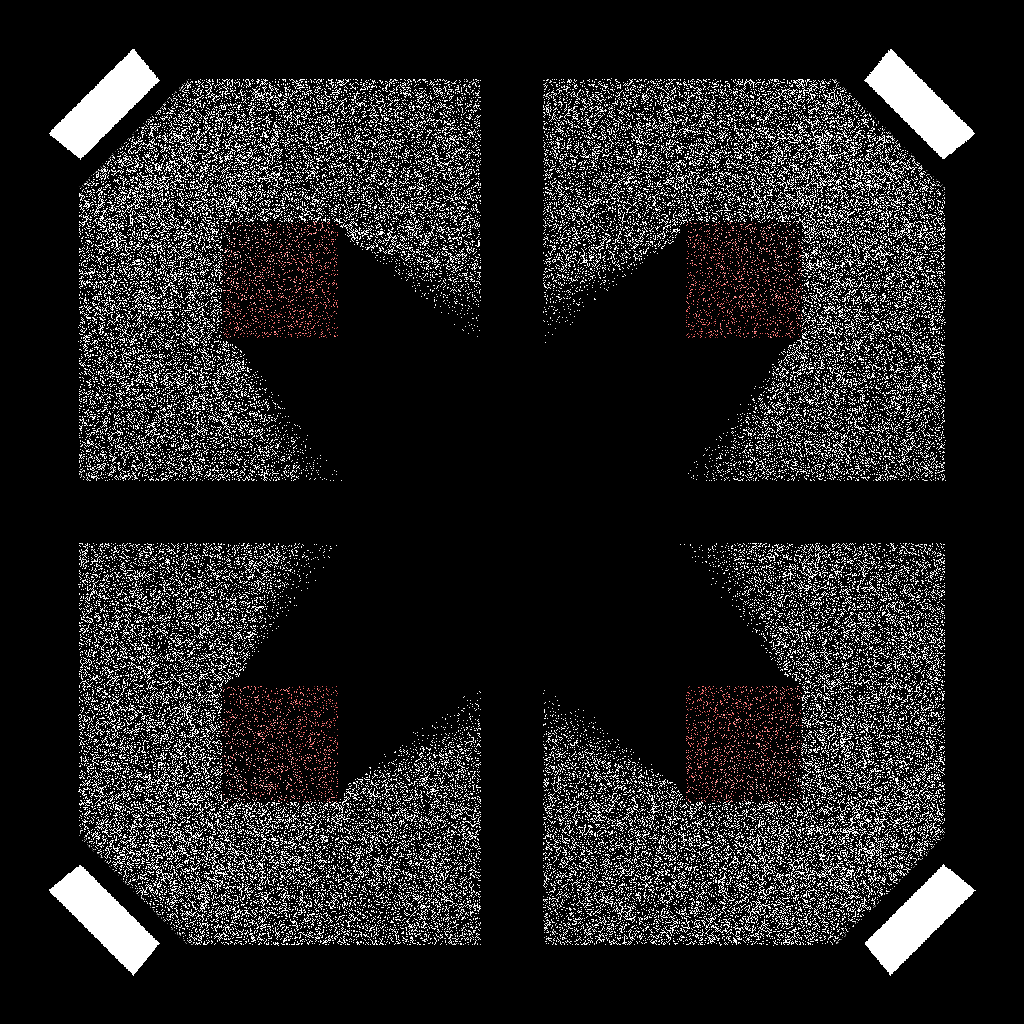}
	\includegraphics[width=0.32\linewidth]{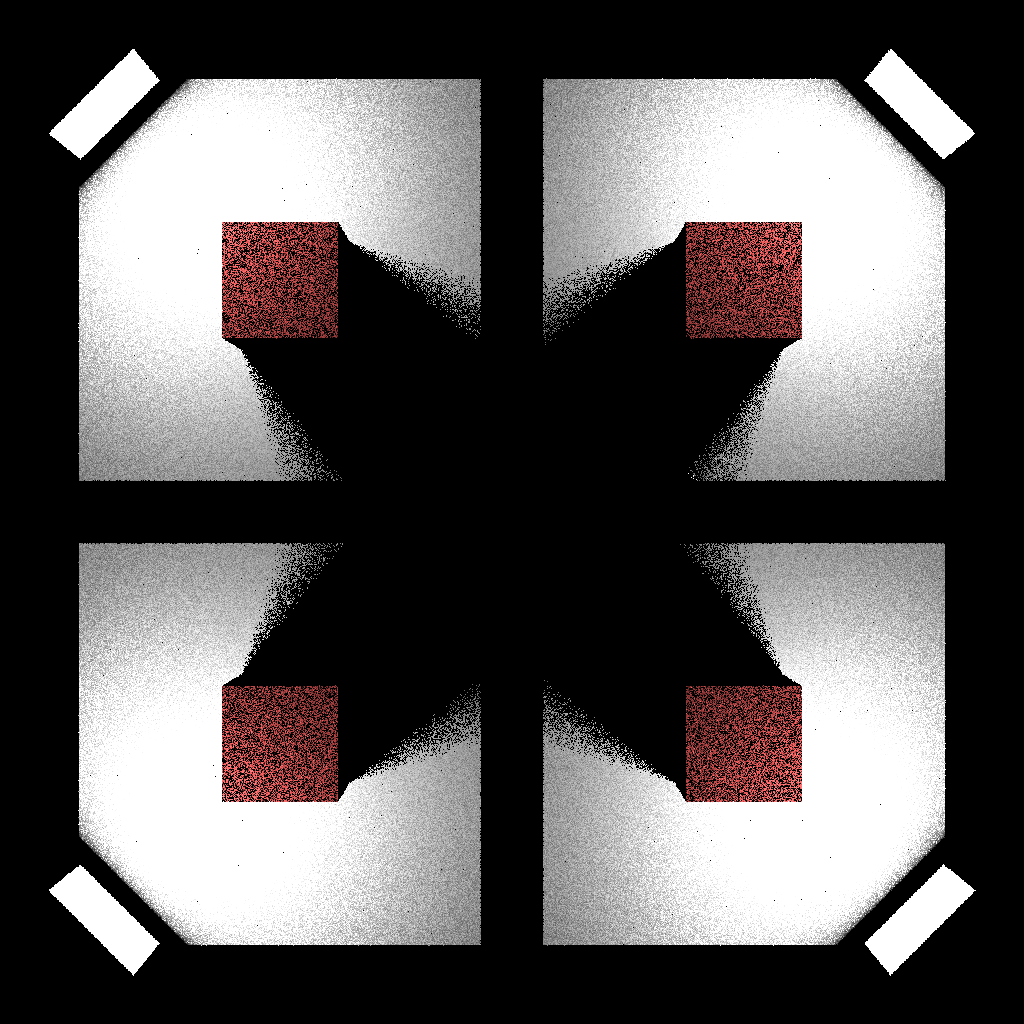}
	\begin{tikzpicture}
\begin{axis}[
xmin = 0, ymin = 0,
legend style={draw=none},
legend style={anchor=north east, fill=none, font=\tiny},
legend cell align=left,
width=.4\linewidth,
height=.4\linewidth]
\addplot[color=gray, name path=f, thick]%
coordinates {
(1, 124.5/255.0)
(2, 113.5/255.0)
(4, 95.3/255.0)
(8, 70.2/255.0)
(16, 46.2/255.0)
(32, 28.9/255.0)
};%
\addlegendentry{random}
\addplot[color=HighlightColor, thick]
coordinates {
(1, 106.5/255.0)
(2, 81.3/255.0)
(4, 52.3/255.0)
(8, 31.2/255.0)
(16, 21.1/255.0)
(32, 13.8/255.0)
};
\addlegendentry{online learning}
\end{axis}
\end{tikzpicture}
\caption{Both images on the left have been rendered at four samples per pixel. To render
the left image we used random sampling for selecting a light source for next event estimation.
The image in the middle was rendered using an artificial neural network trained online
to select a light source with a
high contribution to the point of intersection. The right image shows the relative root
mean squared error (RMSE) of the two methods compared to a ground truth solution where 
the x-axis is the number of samples and the y-axis the RMSE.
\label{Fig:LearnVisibility}}
\end{figure}
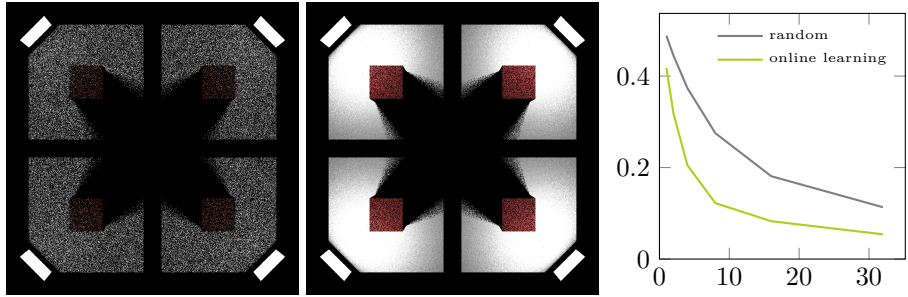

\subsection{Learning Visibility} %

In addition to learning the full light transport and as an alternative to
Sec.~\ref{Sec:TDL}, we can also train an artificial neural
network to approximate the  probability distribution of selecting light
sources with high contribution and thus including visibility. %
During rendering we then feed the intersection details to the
network, which in turn yields a probability distribution over the light
sources. Building and sampling the cumulative distribution function
on-the-fly allows for sampling proportional to this distribution.
Fig.~\ref{Fig:LearnVisibility}
shows the result of this method compared to random selection
of light sources. With just one sample per pixel we can almost perfectly
select a light source. The remaining noise in the 
image is caused by sampling light sources with a finite area. 
The algorithm works similar to the one for learning radiance in Sec.~\ref{Sec:BakingNN}.

\subsection{Discussion}

Using artificial neural networks for high-dimensional approximation
allows for representing both smooth
as well as discontinuous parts of the target function
and capturing view dependent effects such as
for example gloss. Compared
to point-wise representations, less memory is required
and adaptation is automatic through training.

The algorithm in Fig.~\ref{Alg:Rendertraining} (see Sec.~\ref{Sec:OnlineTraining})
samples a density from the artificial neural network from which the cumulative distribution
function is computed. This all can be avoided
by using the artificial neural network to determine the parameters
for an invertible function \cite{DBLP:journals/corr/DinhKB14,DBLP:journals/corr/DinhSB16}.
The invertible function then is used to directly transform a
set of uniform random numbers into a sample according to the learned density.
Published during the revision of our manuscript, these
techniques form the foundation of a new light transport path guiding algorithm
\cite{NeuralImportanceSampling,NeuralImportanceSampling2}. The article represents a big
step forward in importance sampling, as for the first time it becomes possible
to efficiently sample proportional to the solution of an integral
equation.

Common to all approaches is the high cost of evaluating
artificial neural networks, especially, when compared to
techniques affordable in current games. A part of the
performance gap can be addressed by using multiple
small artificial neural networks instead of one large one \cite{GI-ANN}.
Similar to splines, such an approach comes with
competing goals: On the one hand, continuity across the artificial neural
networks has to be established by overlapping
supports. On the other hand, augmenting the
number of input dimensions by what is
called "one-hot encoding" or "one-blob encoding"
\cite{NeuralImportanceSampling}, artificial
neural networks automatically learn an
adaptive representation instead of having to
search for one. Independent
of such choices, all algorithms may be implemented
in so-called primary sample space \cite{GBBE18}, i.e. on the
unit hypercube, rather than in path space.

Our initial experiments have been conducted for static
scenes. Time can be added as another input dimension
to the artificial neural networks as done
in \cite{SkyANN}. However, training an artificial neural
network online in the style of \cite{LightRL,NeuralImportanceSampling,NeuralImportanceSampling2}
is certainly more flexible. Combining the training algorithm
presented in Sec.~\ref{Sec:Train} with the techniques
from \cite{NeuralImportanceSampling,NeuralImportanceSampling2} may lead the path
to real-time reinforcement learning
(see Sec.~\ref{Sec:LightRL} and \cite{LightRL}) not only in computer graphics.

\section{Conclusion}

Building on the close relation of reinforcement learning and
integral equations, a simple algorithm to train
an artificial neural network to approximate the solution of an integral
equation has been derived.
Such a scheme provides a controlled environment that allows one
to explore the smoothness of the progress of learning, suitable initializations of the
artificial neural networks \cite[Sec.6.1]{NeuralImportanceSampling2}, and studying effects of regularization.
In addition, a reference solution always can be computed
and infinitely many training sets can be easily generated.
This allows one to analyze the quality of the approximation
by artificial neural networks and to potentially come up with
a mathematical analysis along the lines of \cite{Approximation-ANN}.

While high-dimensional function approximation in computer graphics becomes
feasible by the application of artificial neural networks, their cost of evaluation
is considerable. Therefore, future research will have to consider the complexity
of artificial neural networks. Besides efficient
algorithms to sample proportional to distributions controlled by
artificial neural networks \cite{DBLP:journals/corr/DinhKB14,DBLP:journals/corr/DinhSB16,NeuralImportanceSampling}, function approximation
by artificial neural networks may improve variance reduction based on the
method of dependent tests, control variates, and the separation
of the main part \cite{Hei98-ML,Keller:99:HMCIS,Rousselle2016CVR}.

\section*{Acknowledgements}

The authors thank Anton Kaplanyan, Thomas M\"uller, and Fabrice Rouselle
for profound discussions and advice.

\end{document}